# A Self-Supervised Framework for Function Learning and Extrapolation

**Simon N. Segert**
Princeton Neuroscience Institute
Princeton University
Princeton,NJ
`ssegert@princeton.edu`

**Jonathan D. Cohen**
Princeton Neuroscience Insitute
Princeton University
Princeton,NJ
`jdc@princeton.edu`

## Abstract

Understanding how agents learn to generalize — and, in particular, to extrapolate — in high-dimensional, naturalistic environments remains a challenge for both machine learning and the study of biological agents. One approach to this has been the use of function learning paradigms, which allow peoples' empirical patterns of generalization for smooth scalar functions to be described precisely. However, to date, such work has not succeeded in identifying mechanisms that acquire the kinds of general purpose representations over which function learning can operate to exhibit the patterns of generalization observed in human empirical studies. Here, we present a framework for how a learner may acquire such representations, that then support generalization — and extrapolation in particular — in a few-shot fashion. Taking inspiration from a classic theory of visual processing, we construct a self-supervised encoder that implements the basic inductive bias of invariance under topological distortions. We show the resulting representations outperform those from other models for unsupervised time series learning in several downstream function learning tasks, including extrapolation.

## 1 Introduction

A key feature of an intelligent agent is the ability to recognize and extend a variety of abstract patterns that commonly occur in the world. We focus on the more tractable but still highly general special case of patterns that take the form of one-dimensional smooth functions. From a formal perspective, the space of all such functions is vast [30], necessitating the use of inductive biases for making useful inferences. Over the past few decades, the empirical studies of function learning in humans has catalogued the forms of several such commonly applied biases, including associative similarity, rule based categorization [26], bias towards positive linear forms [18] and compositional construction from small number of atoms [31]. Taken together, such results describe a class of "intuitive functions", which are functions that people appear readily able to recognize and use.

While it is self-evident that consistent patterns of generalization imply the existence of some previous expectations about the structure of the space of functions, what is not obvious is what mechanisms are capable of learning and encoding such structure through unsupervised or self-supervised experience, in such a way that features relevant to any particular task may be easily "read out" as required. Here we propose a framework for addressing these challenges. The framework consists of two components: a "slow" encoder that learns general purpose representations of one-dimensional functions using a standard self-supervised learning algorithm, and a collection of "fast" heads, which are specialized to different function learning paradigms, and are trained on a small amount of task-specific annotated data using a simple form of supervised learning (linear or logistic regression). The heads are trained on top of the representations learned previously by the encoder, allowing the model to make use of its general knowledge to rapidly adapt to the particular task demands.

Other efforts have taken this general approach [6][11], although none to our knowledge have specifically considered the domain of function learning with intuitive functions – that is, ones that people seem to use. Our approach is further distinguished in the design of the encoder used for self-supervised learning. For this, we treat a scalar function as a (typically very short) time series. The crucial feature of our encoder is a novel family of augmentations of time series, derived from the theory and phenomenology of topological visual processing [42][5]. This theory holds that the visual system is invariant to certain kinds of local topological distortions of stimuli, distortions which we design our augmentations to mimic. Following this idea, we train on a self-supervised objective that tries to enforce invariance across these augmentations, adapting the framework in [6]. We demonstrate that our choice of encoder and training procedure learns representations that perform better on a collection of downstream function learning and generalizaton tasks than do comparison time series representation learning models. This should be of particular interest to the field of semi-supervised learning, since works in that field have not analyzed time series that correspond to intuitive functions.

## 2 Background

### 2.1 Contrastive Learning and InfoNCE

Here we give a brief review of the elements of contrastive learning that are necessary to define our encoder. This is adapted from [6] and [36]. The basic assumption is that we are provided with a set of positive pairs $(v_i, v'_i), i = 1, \ldots, N$, which are interpreted as inputs that we wish to consider similar to each other. All other pairs of inputs are considered as negative pairs, which the objective will attempt to push apart in the latent space. Let $f_\theta : \mathbb{R}^{n_1} \to \mathbb{R}^{n_2}$ and $g_\phi : \mathbb{R}^{n_2} \to S^{n_3-1}$ be two parametric families of functions (e.g. neural networks). Here $n_1$ is the dimensionality of the inputs $v_i$, while $n_2$ and $n_3$ are arbitrary, and $S^{n_3-1}$ denotes the hypersphere consisting of all vectors in $\mathbb{R}^{n_3}$ of unit norm. The objective is

$$\max_{\theta, \phi} \sum_{i=1}^{2N} <(g_\phi \circ f_\theta)(v_i), (g_\phi \circ f_\theta)(v_{i+N})> - LSE^\tau_{j \neq i}(<(g_\phi \circ f_\theta)(v_i), (g_\phi \circ f_\theta)(v_j)>) \quad (1)$$

All indices are interpreted modulo $2N$, where $v_i := v'_{i-N}$ for $i > N$. Furthemore, $\tau > 0$ is a hyperparameter, and $LSE$ denotes the logsumexp function $LSE^\tau_i(z_i) := \tau * \log \sum_i e^{z_i/\tau}$. The brackets $< \cdot, \cdot >$ are the Euclidean dot product. After optimizing this objective, we discard the function $g$ and take the encoder to be the function $f$.

Informally speaking, the first term of the objective function acts to push positive pairs together in the latent space, since it is maximized when both elements in the pair have equal representations. The second term implements the pushing apart of all other pairs of inputs. One way to see this is to use the standard softmax approximation $LSE^\tau_i(z_i) \approx \max_i z_i$. If the the second term in the objective were a hard maximum rather than the LSE, then that term would pick out the negative example in which the representations are most similar, and try to minimize that similarity. Using a softmax instead of a hard maximum ensures that this similarity is in fact being minimized for all negative examples simultaneously to varying degrees. A more precise analysis of properties of this objective may be found in [38].

### 2.2 A Generative model of Intuitive Functions

To define a generative model for reference curves that plausibly resemble the distribution encountered by people, we adapt the generative process proposed in [31]. There, the authors identified a set of 13 Gaussian Process kernels [29] generated from a grammar introduced in [8], each of which is generated by combining a small number of basic structural atoms. This collection of kernels is termed the Compositional Grammar (CG). The authors demonstrate that people learn curves generated from the CG more easily than ones generated from the Spectral Mixture (SM) kernel [40], which is a flexible non-parametric kernel family. Therefore the family of kernels in the CG are good candidates for generating curves that are both naturalistic and are easily reasoned about by people. More precisely, Schulz et. al. define a distribution over intuitive functions that is a mixture model with several different GP kernels. They start with three basic kernels that correspond to basic classes of intuitive functions (namely, linear, sinusoidal, and radial basis kernels), and construct 10 more by taking pointwise sums and products of combinations of these three. We provide further details in an appendix.

It is important to note that, due to the nature of continuous space, in practice it is necessary to represent every function by its values on a finite set $x_1 < \ldots < x_N$ of sample points. In our case, we take the points to be evenly-spaced, and use the same set of points for every function. Thus any function $\{(x_i, y_i)\}$ may be regarded as a time series and vice versa[1]. In what follows we will use the terms "curve", "function" and "time series" interchangeably, with the understanding that the points $x_i$ remain fixed across all functions. Also, since the positions of the $x_i$'s are the same for all functions, we omit them from explicit notation, and use $y$ to denote the vector with components $\{y_i\}_i$ that defines a function.

## 3 A Contrastive Encoder for Intuitive Functions

To define the encoder, following Section 2.1, we need to specify the architecture and the family of positive pairs. For the encoder architecture, we simply take $f$ to be a feedforward network of several 1D convolutions, and $g$ to be an MLP with a single hidden layer. We set $n_2 = n_3 = 128$ and $\tau = .5$. For the class of augmentations, we take inspiration from the field of topological visual perception [5] [42], which posits that the visual system maintains an invariance to local topological distortions (or "tolerances") of stimuli in order to facilitate global processing. In our case, we consider 1-dimensional functions rather than 2-dimensional images, but similar principles apply. We propose a family of transformations that implements localized topological distortions to the function, together with several basic global distortions: (1) random vertical reflection, (2) random jittered upsampling and (3) random rescaling. We denote these respectively by stochastic transformations $T_1, T_2, T_3$, which we describe in more detail below.

The first transformation, that accommodates vertical reflection, is defined by $T_1(y) = -y$ with 50 percent probability and $T_1(y) = y$ otherwise. The second, that accommodates horizontal bending, is the most elaborate. To evaluate $T_2(y)$, we first select a random interval $[a, b] \supset [x_1, x_T]$. We then randomly select points $x'_1 < \ldots < x'_T$ in $[a, b]$. These points are not required to be uniformly spaced. We generate them by sampling uniformly and independently at random from $[a, b]$ and then sorting the samples, and then define $T_2(y)_i = \sum_j C_i e^{-\frac{(x'_j - x_i)^2}{2\sigma^2}} y_j$ where $1/C_i = \sum_j e^{-\frac{(x'_j - x_i)^2}{2\sigma^2}}$. In other words, the values of $T_2(y)$ are given by a Gaussian Kernel Density Estimator (KDE) at the points $x_i$. The effect is three-fold: since the points $x_i$ lie in a proper sub-interval of $[a, b]$, this crops a portion of the function and up-samples to the original resolution. Secondly, because the points $x'_i$ are not evenly spaced, some inhomogeneous horizontal stretching or contraction is introduced. Thirdly, the nature of the Gaussian KDE means that the augmented functions are smoothed with respect to the originals. Finally, we apply $T_3$ that accommodates vertical rescaling. For this, we choose a random interval $[a, b] \subset [0, 1]$ and then apply an affine transformation such that the maximum value of the new function is $b$ and the minimum is $a$. More explicitly, $T_3(y)_i = (b-a)\frac{y_i - \min_j y_j}{\max_j y_j - \min_j y_j} + a$. Since this is applied last, the resulting functions always take values in the interval $[0, 1]$. Therefore the positive pairs take the form $(T_3 T_2 T_1(y), T_3 T_2 T_1(y))$ where $y$ is a function. We reiterate that $T_i$ are stochastic transformations, so despite the notational appearance, the two functions comprising a given positive pair will not be equal, since they are generated using two separate evaluations of a stochastic transformation. In our experiments the function $y$ itself is generated from the Compositional Grammar over intuitive functions described in Section 2.2. An illustration of the augmentations is provided in Figure 3.

## 4 Data description and Encoder training

To evaluate the ability of the encoder to learn a representation of intuitive functions, we generated and trained it on two types of functions: one generated from the family of 13 kernels defined by the CG (see Section 2.2); and the other (used as a control) from a non-compositional SM Kernel, for a total of $14$ kernels. As noted above, [31] showed that human completions are closer to those generated by the CG than by the SM. We included the SM in our generative distribution to allow a similar comparison. Each function was generated by first sampling one of these 14 kernels, then sampling any hyperparameters of that kernel, and finally sampling from the resulting covariance matrix evaluated on $T = 100$ evenly horizontally spaced points. We sampled from the Spectral Mixture kernel 50 percent of the time, and each of the remaining 13 kernels $3.85 (= 50/13)$ percent of the time. Therefore, any differences in the representations between the SM and the CG cannot be ascribed to data availability. We normalized all functions to lie in the interval $[0, 1]$. All encoders were trained using a batch size of 512, with an Adam optimizer with learning rate of .001 and weight decay of $10^{-6}$. All encoders were exposed to 500,000 curves during training. We trained three copies of each encoder using random initializations and averaged results over these copies.

---

[1] In function learning, the x-axis does not necessarily correspond to time. But the same is true of a time series, despite the name: abstractly speaking, it is merely an ordered list of numbers.

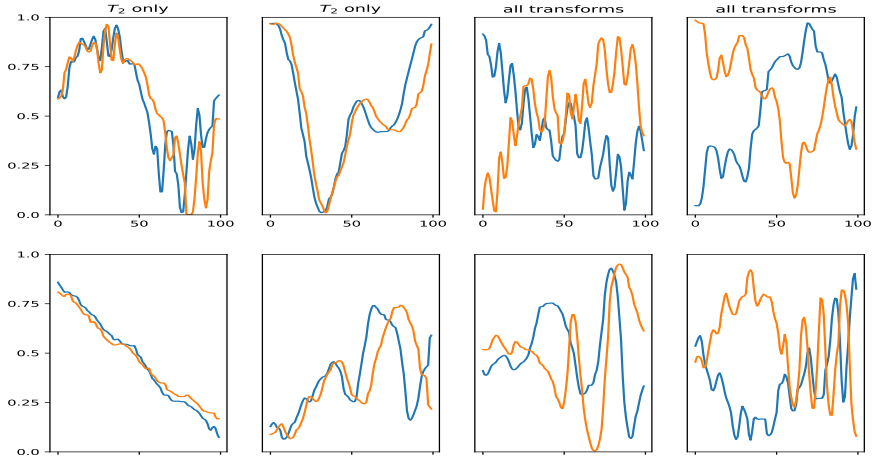

Figure 1: Illustrations of the augmentations. Each plot consists of two functions comprising a positive pair. In the four plots on the left, only the horizontal stretch transformation $T_2$ is applied. In the four plots on the right, all three transformations are applied.

In addition to our Contrastive encoder, for comparison we considered five other encoder modes. Three were unsupervised time series models: Triplet Loss (tloss) [15], Temporal Neighborhood coding (tnc) [34], and Contrastive Predictive Coding (cpc) [36]. We also tested a Variational Autoencoder (vae) [17] as an example of an unsupervised algorithm that has been successful in other domains, but does not exploit any structure particular to time series. To control for the latent space capacity, all encoders had representations of equal dimensionality (128). See appendix for further details on implementation of comparison models. Finally, we also included a baseline encoder ("raw") that simply copies the raw input.

## 5 Results on Downstream Classification and Extrapolation tasks

To evaluate the quality of the learned representations, we adapted three function learning paradigms that are either directly translated from or inspired by paradigms from studies of human performance: (1) kernel classification, (2) multiple choice extrapolation, and (3) freeform extrapolation. The first one corresponds directly to the standard paradigm for unsupervised learning evaluation in computer vision [6], in which an unsupervised algorithm is trained on a dataset for which ground truth annotations are available, and then a supervised classifier such as a logistic regression is fit on top of the frozen representations. Although to our knowledge this has not been used directly in the analysis of empirical results concerning human function learning, it may be regarded as an abstraction of the experiments from [19], which showed that people's completions depend on their judgements about the category to which a function belongs, suggesting that people make use of categories when judging functions. The two extrapolation tasks are drawn directly from [31]. A version of the third task also appears in[41], however using a different generative process for the probe curves.

Table 1: Accuracy on the categorization task, as a function of the number of training examples per category. Chance performance is 7.14 percent

|  | 3 | 10 | 30 | 100 | 300 |
|---|---|---|---|---|---|
| contrastive (ours) | **40.95± 1.83** | **55.27± 1.41** | **64.81± 1.80** | **72.00± 1.52** | **76.23± 1.31** |
| cpc | 25.06± 1.29 | 35.45± 1.57 | 46.38± 1.14 | 54.94± 1.18 | 58.48± 1.04 |
| raw | 11.86± 1.47 | 15.54± 2.02 | 14.27± 1.96 | 15.73± 1.51 | 14.25± 1.64 |
| t-loss | 30.41± 1.73 | 41.31± 1.89 | 52.16± 1.20 | 59.78± 1.24 | 63.35± 1.19 |
| tnc | 23.15± 1.14 | 31.22± 1.09 | 38.55± 1.23 | 44.85± 1.08 | 49.16± 1.20 |
| vae | 9.27± 1.13 | 12.77± 1.65 | 21.51± 1.89 | 29.17± 1.56 | 33.74± 1.50 |

For each task, we define a head that transforms the encoder representations to a task-specific output, and train the head on a small amount of labeled data. In all cases when training the heads, the weights of the encoder are frozen.

In addition, we note that several of the tasks required that we compute the posterior mean with respect to a Gaussian kernel in order to construct the training data, which required that we fit initially the kernel hyperparameters. For example, in the multiple choice task, to construct two candidate completions we took the posterior mean of the prompt curve with respect to both the SM and the CG kernels, which required we first fit the hyperparameters for those kernels. The fitting of GP kernel hyperparameters is known to suffer from under-fitting and instability problems (see [41], including the supplementary material). To address this, we fixed the hyperparameter values of each kernel class, and evaluated the downstream performance of all encoders on curves generated with this fixed set of hyperparameters. We repeated this 10 times using different random choices of hyperparameters each time, and averaged the results. This ensured that the hyperparameter values were correctly specified within each task, and that our results were not influenced by the imperfections of any particular hyperparameter optimization procedure.

We trained three copies of each of the six encoders (contrastive encoder and five comparisons) using different initializations, and all reported results were averaged over the 3 copies and 10 hyperparameter choices, for a total of 30 measurements. The error bars are 95 percent confidence intervals of the standard error of the mean over those measurements.

### 5.1 Kernel classification

Here, the task was to predict which of the 14 kernels was used to generate a given function. The head was simply a linear layer + softmax, where the outputs were interpreted as the probabilities of each class. Thus it is equivalent to a 14-way logistic regression on the encoder representations. In all cases, we fit the head using the SGDClassifier class from scikit-learn. Additionally, we separately chose an L2 penalty for each encoder using cross validation. As shown in Table 1, the contrastive encoder is able to attain approximately 55 percent accuracy using only 10 labeled examples per class, which improves to approximately 75 percent when using 300 examples per class, improving upon the second-best model (t-loss) by around 10 percentage points.

### 5.2 Multiple Choice Extrapolation

In the multiple choice completion paradigm, the models were presented with a prompt curve $y \in \mathbb{R}^{80}$, as well as several candidate completions curves $y^i \in \mathbb{R}^{100}$ with the properties that $y_j^i = y_j$ for $j <= 80$ and were required to select the correct completion. Following Schulz et. al., we constructed the candidate completion curves by computing the posterior mean with respect either to the SM kernel, or to the CG kernel family, with the correct answer corresponding to which of these two kernels was used to generate the prompt curve. The training data for the head consisted of 50 percent prompt curves sampled from the SM and 50 percent curves sampled from the CG.

Since this task required comparing the prompt curve to each of the candidate curves, we used a quadratic decision rule for the head. Let $h_0$ denote the encoder representation of the prompt (upsampled to 100 points prior to being fed into the encoder), and $h_i, i > 0$ denote the representations of the choices. The head linearly projected these vectors into a lower-dimensional space, and chose between the alternatives using a dot product in this space. That is, we fit a model of the form $p_i \propto e^{(wh_i, wh_0)}$ where $w$ is a linear projection from the encoder space to $\mathbb{R}^{32}$ and $p_i, i = 1, 2$ are the choice probabilities. All heads were trained on a cross-entropy loss using the Adam optimizer with a learning rate of .01.. In this case, we see from Table 2 an improvement in accuracy of approximately 5 to 10 percent compared with the second best model. Interestingly, the rank ordering of the models also differs compared to the categorization task: here cpc and tnc both outperform t-loss, and the vae generally matches performance of the raw encoder baseline.

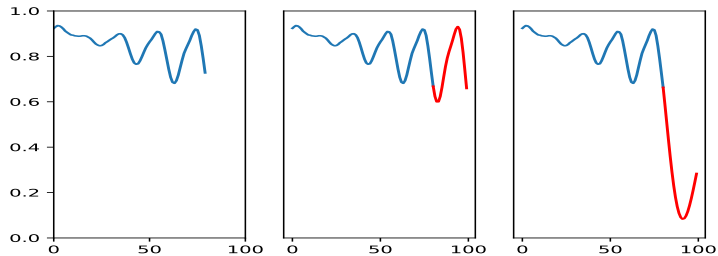

Figure 2: An example multiple choice completion problem. The prompt curve is on the left. The compositional completion is in the middle and the mixture completion is on the right. In this case, the correct answer is the compositional completion. The coloring of the candidate curves is for visual aid only.

Table 2: Performance on the Multiple Choice completion task, as a function of the number of training examples per category. Chance performance is 50 percent.

|  | 3 | 10 | 30 | 100 | 300 |
| --- | --- | --- | --- | --- | --- |
| contrastive (ours) | **67.74± 2.48** | **73.68± 1.69** | **78.09± 1.47** | **79.07± 1.75** | **80.90± 1.93** |
| cpc | 62.80± 2.60 | 67.55± 2.26 | 70.55± 1.93 | 71.58± 1.97 | 74.48± 1.81 |
| raw | 51.64± 2.83 | 54.80± 2.13 | 54.28± 2.20 | 54.68± 1.85 | 56.96± 2.22 |
| t-loss | 60.18± 1.44 | 62.74± 1.62 | 65.12± 1.57 | 65.31± 1.55 | 67.76± 1.73 |
| tnc | 58.38± 2.46 | 63.65± 1.89 | 68.91± 1.64 | 70.05± 1.68 | 72.25± 1.99 |
| vae | 52.48± 0.74 | 53.03± 0.83 | 53.32± 0.85 | 53.57± 0.84 | 55.00± 1.29 |

### 5.3 Freeform extrapolation

In this task, for a given function $y \in \mathbb{R}^{100}$, the model was presented with an initial portion $y_{1:80}$ and required to make a prediction $\hat{y} \in \mathbb{R}^{20}$ that extended it for a fixed sized window (length 20). Performance was measured by $Sim(\hat{y}, y_{80:100})$, for some choice $Sim$ of similarity function. It has been argued that, due to its high-dimensional and underconstrained nature, this task provides a more rigorous test of extrapolation that do discrete catorization tasks and that, in an empirical setting, it may provide finer-grained insights into peoples' inductive biases [7]. However, it may be unreasonable to expect that an algorithm trained without *any* predictive experience can exhibit a reasonable ability to free-form extrapolate. Here, we tested the hypothesis that this capacity can arise from modest amounts of supervised (predictive) training that based on categorization judgements among functions representations acquired in a self-supervised from by the contrastive learning mechanism described above. To test this, we implemented a simple form of curriculum learning. In the first phase of the curriculum, we trained a logistic regressor on the encoder features learned during unsupervised training, to predict the generative kernel of an input function, exactly as in the categorization task from Section 5.1. Here we presented the regressors with 300 functions and corresponding category annotations from each kernel. In the second phase of the curriculum, we present a small number of functions $y^k$ with no label annotations. We then fit a simple class-dependent forecasting model of the form:

$$y_i^k \sim w_0^{\hat{c_k}} + w_0 + \sum_{j=1}^{L}(w_j^{\hat{c_k}} + w_j)y_{i-j}^k \quad (2)$$

where $\hat{c}_k \in \{1, 2, \ldots, 14\}$ denotes the kernel class of the function $y^k$ predicted by the logistic regressor. Here $L$ is a hyperparameter that controls the autoregressive time lag. We set $L = 20$ in all cases. The parameters $\{w_j^m\}_{0 \leq j \leq L, 1 \leq m \leq 14}$ are weights that are fit using least-squares. When given a function $y$ to extrapolate at test, we first estimated the class $\hat{c} \in \{1, 2, \ldots, 14\}$ of $y$ using the logistic regressor and encoder features. We then forecast it using the autoregression weights $\{w_j + w_j^{\hat{c}}\}_{0 \leq j \leq L}$.

We compared the results of this procedure with two controls. The first was an Ideal Observer model that was given access to the true underlying Gaussian kernel used to generate each function, information to which the other models were not privy. This model forecast a given function by computing the posterior mean with respect to the kernel on which it was trained. Since this model used Bayesian inference on the exact underlying distribution over functions, it represented the best performance that any model could attain. We refer to this as the "GPIO" (Gaussian process Ideal Observer) model. The second control removed the categorization step, and so measured the contribution of the categorization to the forecast quality. It used an unconditional forecasting model of the form $y_i^k \sim w_0 + \sum_{j=1}^{L} w_j y_{i-j}^k$ that ignored any category structure. This model was trained on exactly the same number of functions as the other forecasting models (with the number of curves used to train the logistic regressor included in this count).

We evaluated the extrapolation performance of each model using the Pearson correlation coefficient and L2 distance (see appendix for results of L2 distance) between the actual and predicted curves. The results, shown in Table 3, are similar to those for the categorization task. All models substantially outperformed the autoregression baseline, indicating that even imperfect category information is helpful for extrapolation. The contrastive model performed better than any other model except the GPIO model. Several example extrapolations from the contrastive model are shown in Figure 3.

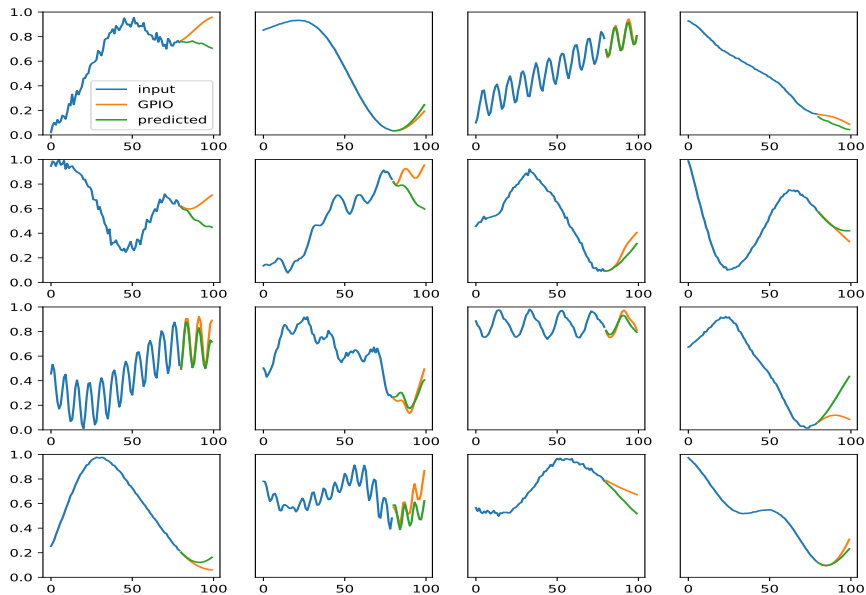

Figure 3: Freeform completions generated by the contrastive model, using the maximal amount of training data. The GPIO completions are also shown for comparison.

## 6 Related Work

### 6.1 Contrastive Learning and Time Series Representation Learning

The idea of learning representations by maximizing an information theoretic criterion can be traced at least back to Linsker's InfoMAX [21], in which it was shown that certain properties of neurons in visual cortex could be replicated by training the encoder to maximize mutual information between the input and the encoder representation. This principle was subsequently extended to the problem of unsupervised deconvolution of time series [4] by extraction of independent components. A network that learns by instead trying to maximize representational similarity between two different parts of the same input, presaging the modern approach to contrastive learning, was introduced in [3]. Rather than trying to optimize the mutual information directly, however, most modern implementations of this idea use instead a form of the so-called InfoNCE loss, introduced in [36]. There it was shown that this objective is a tractable lower bound to the mutual information criterion, which can be difficult to estimate directly. This loss is also strongly reminiscent of the older technique of Contrastive Hebbian learning [12], insofar as both involve computing average network activations over a set of "positive pairs" of inputs as well as over a set of "negative pairs", and trying to maximizing the difference between the two averages. The authors incorporated this objective in their Contrastive Predictive Coding model (CPC), in which a recurrent encoder is trained to predict its own future outputs. This basic loss function has been adapted and modified in several ways. In [6] it is used in tandem with a siamese network architecture, as we described in Section 2.1, while [1] extends this setup to a seq-to-seq objective and [20] integrates the contrastive objective with a reconstructive one. A similar contrastive objective is used in [11], except with a memory bank used to sample negative examples, with this approach extended to videos in [28]. The approach of [14] is also very similar in spirit, in which the idea is to learn temporal features of time series that differ across different time windows.

Table 3: Results on the freeform task, as a function of the number of functions per category use to train the regression (not counting the functions used to train the kernel classifier). Values are the Pearson correlation of the predicted to the true completion. The value for the GPIO model is 83.70± 1.78.

|  | 1 | 3 | 10 | 30 | 100 |
|---|---|---|---|---|---|
| autoregression | 18.48± 4.46 | 18.48± 4.48 | 18.47± 4.52 | 18.50± 4.47 | 18.66± 4.47 |
| contrastive (ours) | **30.15± 2.36** | **49.05± 2.00** | **60.78± 1.59** | **63.34± 1.39** | **63.91± 1.43** |
| cpc | 25.13± 2.83 | 42.32± 2.61 | 51.04± 2.39 | 53.06± 1.92 | 54.79± 1.92 |
| raw | 19.89± 4.20 | 23.13± 4.81 | 27.02± 5.82 | 28.18± 5.81 | 29.14± 5.89 |
| t-loss | **29.45± 3.28** | 44.48± 2.74 | 54.62± 2.00 | 56.87± 1.93 | 58.10± 1.95 |
| tnc | **26.72± 3.22** | 39.11± 2.84 | 48.30± 1.81 | 51.52± 1.60 | 52.78± 1.74 |
| vae | **27.36± 3.09** | 33.51± 2.53 | 37.18± 2.67 | 39.27± 2.51 | 40.89± 2.39 |

Although some of the works above deal with sequential data, they tend to be high-dimensional (videos [28], image patches [1]) or otherwise not directly human interpretable (audio [36], radio frequency signals [20]). Fewer works consider the more specific case of representation learning of 1-dimensional time series, which are the sorts of stimuli used in function learning experiments. Two particularly notable models here are Triplet loss [15] and Temporal Neighborhood Coding [34]. The first is inspired by word2vec [27], and relies on predicting the representation of a "word" (here a short window of the time series) from the representation of its "context" (here a longer window containing the "word"). In TNC, the timeseries is divided into disjoint segments. The encoder is jointly learned alongside a discriminator, in such a way that the discriminator is able to tell the difference between distant and proximal observations. In both TNC and Triplet loss, a recurrent encoder is used. An alternative approach to unsupervised learning of 1D time series learning is through autoencoders. A popular choice here is a seq2seq architecture with a reconstruction loss [2] [23] [25]. [24] augmented this setup with a k-means objective to encourage clustering in the latent space. Compared to our approach, these involve considerably more complexity, through the use of an additional decoding step, as well as more intricate seq2seq architectures.

### 6.2 Function Learning and Gaussian Processes

The dominant framework for modeling of function learning is through Gaussian processes, a statistical model which specifies a probability distribution over the infinite-dimensional space of functions and allows for tractable inference procedures [29]. Lucas et. al. [22] used Gaussian processes to capture a wide range of empirical function learning phenomena, while [41] and [31] proposed specific families of kernels to model human extrapolation judgements. A limitation of the basic GP framework is its dependence on a choice of specific kernels or kernel families. Our approach sought to avoid this dependence through unsupervised learning. Other approaches have taken a similar tack. The Spectral Mixture Kernel [40] and Variational GP [35] do so by introducing nonparametric families of kernels that can approximate arbitrary kernels to an arbitrary level of precision. [8] implement a similar idea, except by building up a family of kernels using operations of a small number of atomic kernels, and performing a search over the resulting combinatorial space. [32] perform a similar search except using a continuous relaxation and neural network. In [13], an appropriate kernel is found by fitting a Boltzmann machine. Neural Processes go further and [9] replace the Gaussian kernel with a more flexible parametric family of distributions that can be learned using a neural network. Such approaches share our broad goal of trying to learn the structure of a space of curves without assuming any particular functional forms ahead of time. But they differ from ours in that they both build in more statistical machinery and do not learn internal representations.

## 7 Limitations

There are several noteworthy limitations to our encoder. First, it differs from the other models in that it was not designed to necessarily scale to very long time series. In particular, we use a feedforward convolutional encoder that processes the entire time series at once, while the other techniques use some combination of recurrence and/or local windowing of the time series. Our time series have only 100 points, which is extremely short from the viewpoint of typical time series learning models. However, in the context of function learning, such short time series have face validity, because during function learning experiments people can only make use of limited information at a time [37]. Thus, while it is not clear how well our encoder architecture would scale to very long time series, it is also not clear how well humans would do so either; and it remains to be determined how useful doing so would be for generalization in natural environments. These remain subjects for future research. Second, the feedforward nature of our encoder also restricts it to processing time series of a fixed length, as opposed to the recurrent encoders of the other models which can handle time series of variable lengths. In principle this could be overcome by upsampling or downsampling as necessary (and this was the approach we took in the multiple choice completion task). While this kind of resampling may be benign or helpful in certain circumstances (e.g., as a form of context normalization [39]), there are also many applications in which it would instead be preferable to preserve the original resolution and accommodate variable lengths.

## 8 Discussion

The contrastive encoder we presented exhibited superior performance to comparison models in tests of generalization involving categorization as well as free form extrapolation. This was the case, despite its greater simplicity than those models (other than the baseline models). For testing free form extrapolation, we used a curriculum learning strategy that involved first learning categories and then learning category-specific forecasting rules. While this procedure was more complex than for the other heads, there is reason to believe that it in fact resembles the process by which people may learn to make inferences in sparse and undetermined settings. The most direct evidence of this in the realm of function learning comes from [19], which showed that peoples' extrapolations of curves were dependent on whether they judged the curves to lie in a previously encountered category, suggesting that people use category-dependent forecasting rules. More generally, our approach may be viewed as implementing a form of a Hierarchical Bayesian model, which have been show to capture the structure of peoples' intuitive theories about abstract structures in the world [10][33][16].

Our approach also fits with the idea that learning in natural agents involves adjudicating a tension between maintaining as much flexibility as possible (by optimizing a Maximum Entropy objective) while at the same time maximizing efficiency of computation (e.g., by optimizing a Minimum Energy objective) . From this perspective, the contrastive encoder can be viewed as maximizing entropy, as implemented by the InfoNCE objective that we used, as it may be shown [38] that the second term in that objective is an estimator of the entropy of the distribution of codes in the latent space. Complementing this, our curriculum learning can be viewed as minimizing the energy of representations generated by a given category of function when presented with an instance of that function. This may strike a balance between flexibility (of generalization) and efficiency (of inference) that begins to approximate the balance observed in natural agents, and humans in particular.

# 9  Acknowledgements

Thanks to Taylor Webb and Steven Frankland for comments on the manuscript, and to Zack Dulberg and Sebastian Musslick for helpful discussions. S.N.S is supported by a T32 Training Grant in Computational Neuroscience (T32MH065214). This project was made possible through the support of a grant from the John Templeton Foundation. The opinions expressed in this publication are those of the authors and do not necessarily reflect the views of the John Templeton Foundation.

[21] R. Linsker. Self-organization in a perceptual network. *IEEE Computer*, 1988.

[22] C. G. Lucas, T. L. Griffiths, J. J. Williams, and M. L. Kalish. A rational model of function learning. *Psychonomic Bulletin Review*, 2015.

[23] X. Lyu, M. Hueser, S. L. Hyland, G. Zerveas, and G. Rätsch. Improving clinical predictions through unsupervised time series representation learning. *arXiv:1812.00490*, 2018.

[24] Q. Ma, J. Zheng, S. Li, and G. W. Cottrell. Learning representations for time series clustering. In *Advances in Neural Information Processing Systems*, 2019.

[25] P. Malhotra, L. V. Vishnu TV, P. Agarwal, and G. Shroff. Timenet:pre-trained deep recurrent neural network for time series classification. *arXiv:1706.08838*, 2017.

[26] M. A. McDaniel and J. R. Busemeyer. The conceptual basis of function learning and extrapolation: Comparison of rule-based and associative-based models. *Psychonomic Bulletin Review*, 2005.

[27] T. Mikolov, I. Sutskever, K. Chen, G. Corrado, and J. Dean. Distributed representations of words and phrases and their compositionality. In *Advances in Neural Information Processing Systems*, 2013.

[28] T. Pan, Y. Song, T. Yang, W. Jiang, and W. Liu. Videomoco: Contrastive video representation learning with temporally adversarial examples. In *Computer Vision and Pattern Recognition*, 2021.

[29] C. E. Rasmussen and C. K. I. Williams. *Gaussian processes for machine learning*. MIT Press, 2006.

[30] M. Reed and B. Simon. *Functional analysis*. Academic Press, 1980.

[31] E. Schulz, J. B. Tenenbaum, D. Duvenaud, M. Speekenbrink, and S. J. Gershman. Compositional inductive biases in function learning. *Cognitive Psychology*, 2017.

[32] S. Sun, G. Zhang, C. Wang, W. Zeng, J. Li, and R. Grosse. Differentiable compositional kernel learning for gaussian processes. *International Conference on Machine Learning*, 2018.

[33] J. B. Tenenbaum, C. Kemp, T. L. Griffiths, and N. D. Goodman. How to grow a mind: Statistics,structure, and abstraction. *Science*, 2011.

[34] S. Tonekaboni, D. Eytan, and A. Goldengerg. Unsupervised representation learning for time series with temporal neighborhood coding. In *International Conference on Learning Representations*, 2021.

[35] D. Tran, R. Ranganath, and D. M. Blei. The variational gaussian process. In *International Conference on Learning Representations*, 2016.

[36] A. van den Oord, Y. Li, and O. Vinyals. Representation learning with contrastive predictive coding. *arXiv:1807.03748*, 2018.

[37] P. L. Villagra, I. Preda, and C. G. Lucas. Data availability and function extrapolation. In *Proceedings of the Cognitive Science Society*, 2018.

[38] T. Wang and P. Isola. Understanding contrastive representation learning through alignment and uniformity on the hypersphere. In *International Conference on Machine Learning*, 2020.

[39] T. W. Webb, Z. Dulberg, S. M. Frankland, A. A. Petrov, R. C. O'Reilly, and J. D. Cohen. Learning representations that support extrapolation. In *International Conference on Machine learning*, 2020.

[40] A. G. Wilson and R. P. Adams. Gaussian process kernels for pattern discovery and extrapolation. In *International Conference on Machine Learning*, 2013.

[41] A. G. Wilson, C. Dann, C. G. Lucas, and E. P. Xing. The human kernel. In *Conference on Neural Information Processing Systems*, 2015.

[42] E. Zeeman. Topology of the brain. In *Mathematics and computer science in biology and medicine: Proceedings of the conference held by the Medical Research Council in Association with the Health Department*, 1965.

## Appendix A  Code Availability

We provide code and pretrained models at: `https://github.com/SimonSegert/functionlearning-contrastive`

## Appendix B  Encoder architecture

Let Linear(m) denote a linear layer with an output size of $m$. Further, let 1dConv(c,k,s) denote a 1-dimensional convolution layer with a filter size of $k$, a stride of $s$ and $c$ output channels, and let 1dMaxPool(k) denote a 1-dimensional max pooling with kernel size of $k$. Then our encoder is given by the following sequence of layers:

```
1dConv(64,5,2)
1dMaxPool(2)
LeakyReLU()
BatchNorm()
1dConv(64,5,1)
1dMaxPool(2)
LeakyReLU()
BatchNorm()
1dConv(64,3,1)
LeakyReLU()
Linear(128)
```

## Appendix C  Comparison models

For t-loss [15] and tnc [34] we used the implementations made available on the authors' Github pages. Since the authors of cpc [36] do not provide an official public implementation, we used the cpc implementation provided in the tnc repo. The t-loss repository is licensed under Apache License 2.0 and the tnc repository is not licensed. For the vae, we wrote our own implementation using the same encoder architecture described in the previous section. The decoder consisted of the same pieces in the opposite order, except with the convolutional layers replaced with ConvolutionTranspose layers, and the MaxPooling layers replaced with MaxUnpooling layers. Also, we do not use batch normalization in the decoder.

Both tnc and cpc require choice of a window size, which as the authors of tnc note [34], must in general be chosen using domain knowledge of the time series in question. In our case, the time series are very smooth, meaning that small windows will generally fail to contain distinguishing features. For example, if the time series contains a slow linear trend (which many curves from the CG do), this may be hard to detect in a small window. Thus for both of these models we chose the window size to be the largest divisor of 100 (the number of observations in each time series) that could be accommodated by the implementation.

For t-loss and cpc, extracting a global representation of a time series is straightforward, and we did so as described in the respective papers. For tnc, a key aspect of the model is that it learns representations of windows of the time series, rather than a representation of the full time series itself, which allows it to deal with non-stationarity. To convert this to a global representation, we divided the time series into disjoint windows and concatenated the window representations. This was the representation used in all of our downstream analyses. We also tried a variant in which the window representations were averaged rather than concatenated, but found that the results were slightly worse.

As described in the main text, we modified all comparison models so that the resulting representations had equal dimensionality of 128 (for tnc this refers to the global representation formed by concatenating window representations). Other than this modification, and the choice of window size described above, we used the hyperparameter settings provided in the implementations.

## Appendix D  Further Description of Compositional Grammar and Spectral Mixture Kernels

Here we review the set of kernels used by Schulz et. al. [31] as well as the spectral mixture kernel [40]. We also specify our choice of kernel hyperparameter distributions, which were used to sample curves as described in the main text.

We begin with a brief refresher on Gaussian process formalism, and refer to [29] for further details. Recall that a *kernel function* is a function $K : \mathbb{R} \times \mathbb{R} \to \mathbb{R}$, which may be intuitively thought of as an infinite-dimensional covariance matrix. More precisely, a Gaussian Process with kernel $K$ is a distribution over the space of all functions $f : \mathbb{R} \to \mathbb{R}$ with the property that for any finite set of points $\{x_i\}_{i=1}^n \subset \mathbb{R}$, the vector $(f(x_1), \ldots, f(x_n))$ has a multivariate normal distribution with mean 0 and covariance matrix $C_{ij} = K(x_i, x_j)$. The kernel function is required to be such that this matrix is always symmetric and non-negative definite.

In our case, the points $x_i$ are always fixed to be 100 evenly spaced points between 0 and 10. Thus the Gaussian process formalism is equivalent to a multivariate normal distribution, but we will retian the language of Gaussian processes for consistency with Schulz et. al.

A Gaussian process kernel can be used to predict future values of a curve. Suppose that we are given some partial set of observations $y_{obs} := \{y_i\}_{i=1}^m$ where $m < n$, and we want to estimate the values $y_{ext} := \{y_i\}_{i=m+1}^n$ on the remaining points. It may be shown [29] that the posterior mean estimate is given by

$$\mathbb{E}(y_{ext}|y_{obs}, K) = K(x_{m+1:n}, x_{1:m}) K(x_{1:m}, x_{1:m})^{-1} y_{obs} \qquad (3)$$

We used this to formula to generate completions in the Multiple Choice Completion task and the Freeform Completion task.

Schulz et. al. define the Compositional Grammar by starting from three basic kernels:

$$\begin{aligned}
K_{linear}(x_i, x_j) &= (x_i - \theta_1)(x_j - \theta_1) \\
K_{rbf}(x_i, x_j) &= \theta_3 e^{-(x_i - x_j)^2/\theta_2^2} \\
K_{periodic}(x_i, x_j) &= \theta_4 e^{-\sin^2(2\pi|x_i - x_j|\theta_5)/\theta_6^2}
\end{aligned}$$

where $\theta_i$ are hyperparameters. In addition to these three, the authors take advantage of the algebraic fact that the pointwise sum and product of two kernels are themselves valid kernel functions. The remaining kernels in the CG are given by

$$\begin{aligned}
K_{linear} &+ K_{periodic} \\
K_{linear} &+ K_{rbf} \\
K_{rbf} &+ K_{periodic} \\
K_{linear} &* K_{periodic} \\
K_{linear} &* K_{rbf} \\
K_{rbf} &* K_{periodic} \\
K_{linear} &+ K_{rbf} + K_{periodic} \\
K_{linear} &+ K_{periodic} * K_{rbf} \\
K_{periodic} &+ K_{linear} * K_{rbf} \\
K_{linear} &* K_{rbf} * K_{periodic}
\end{aligned}$$

The SM kernel[40] is defined by the formula

$$K_{mix}(x_i, x_j) = \sum_{i=1}^m w_i e^{-2\pi^2 (x_i - x_j)^2 \sigma_i} \cos(2\pi (x_i - x_j) \mu_i)$$

In our experiments, we sampled the hyperparameters of each of these kernel families at random. To specify these hyperparameter distributions, we use the convention that $N(a, b)$ denotes a normal distribution with mean $a$ and standard deviation $b$. Also $U[a, b]$ denotes a uniform distribution between $a$ and $b$, and $U(\{z_1, \ldots, z_k\})$ denotes a categorical uniform distribution on the finite set $\{z_1, \ldots, z_k\}$. The distributions over kernel hyperparameters are:

$$\begin{aligned}
\theta_1 &\sim N(0, 2) \\
\theta_2 &\sim U[1, 5] \\
\theta_3 &\sim U[1, 3] \\
\theta_4 &\sim U[1, 3] \\
\theta_5 &\sim U[0, .5] \\
\theta_6 &\sim U[1, 5] \\
\mu_i &\sim N(0, .01) \\
\sigma_1 &\sim N(0, .02) \\
w_i &\sim U[0, 1] \\
m &\sim Unif(\{2, 3, 4, 5, 6\})
\end{aligned}$$

Table 4: Similar to Table 3 in the main text, except using L2 distance instead of Pearson correlation. The value for the GPIO is $.0561 \pm .0082$.

|  | 1 | 3 | 10 | 30 | 100 |
|---|---|---|---|---|---|
| ar | **0.1449± 0.0165** | 0.1450± 0.0165 | 0.1451± 0.0165 | 0.1451± 0.0164 | 0.1453± 0.0164 |
| contrastive (ours) | 0.2058± 0.0101 | **0.1326± 0.0085** | **0.0989± 0.0062** | **0.0945± 0.0058** | **0.0926± 0.0057** |
| cpc | 0.2223± 0.0164 | 0.1444± 0.0073 | 0.1143± 0.0064 | 0.1108± 0.0065 | 0.1086± 0.0066 |
| raw | 0.1758± 0.0201 | 0.1562± 0.0173 | 0.1423± 0.0160 | 0.1384± 0.0151 | 0.1367± 0.0154 |
| t-loss | 0.2044± 0.0120 | **0.1374± 0.0078** | 0.1085± 0.0056 | 0.1037± 0.0058 | 0.1017± 0.0058 |
| tnc | 0.2019± 0.0131 | 0.1453± 0.0064 | 0.1199± 0.0066 | 0.1147± 0.0066 | 0.1131± 0.0066 |
| vae | 0.1921± 0.0128 | 0.1504± 0.0081 | 0.1330± 0.0072 | 0.1287± 0.0074 | 0.1273± 0.0076 |

## Appendix E  Further details on augmentations

When constructing the augmentation $T_2$, we choose the random interval $[a, b] \supset [x_1, x_T]$ such that $a > x_1 - .4 * (x_T - x_1)$ and $b < x_T + .4(x_T - x_1)$. The bandwidth parameter in the Gaussian KDE is set to $\sigma = .1$. When constructing the augmentation $T_3$, we define the random interval $[a, b] \subset [0, 1]$ such that $b - a > .8$.

## Appendix F  Illustration of categorization task

We display several curves and their corresponding category annotations in Figure 4.

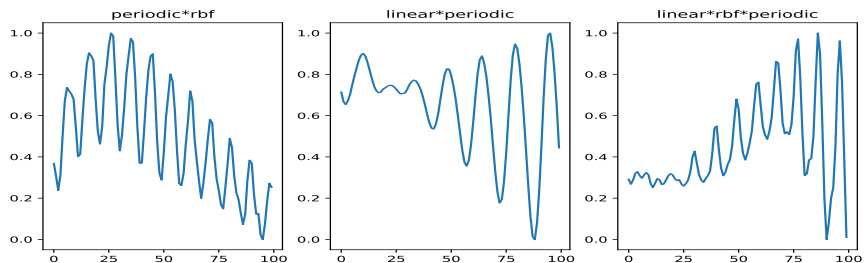

Figure 4: Several examples of annotated data used for training on the Kernel Classification task.

Table 5: Value of $\Delta_{acc}$ on the multiple choice task

|             | 3            | 10           | 30           | 100          | 300          |
|-------------|--------------|--------------|--------------|--------------|--------------|
| contrastive | 19.82± 5.88  | 21.49± 5.26  | 20.53± 4.87  | 18.57± 4.84  | 19.39± 4.81  |
| cpc         | -3.92± 7.59  | 1.50± 5.50   | 4.73± 5.60   | 6.38± 4.43   | 10.33± 4.02  |
| raw         | 5.19± 5.36   | 2.25± 3.79   | 3.87± 2.45   | 5.09± 1.81   | 7.92± 2.61   |
| t-loss      | 0.57± 5.04   | 5.46± 4.37   | 8.24± 3.95   | 10.30± 3.44  | 12.72± 3.75  |
| tnc         | -3.23± 6.13  | 4.70± 4.07   | 6.42± 2.81   | 8.18± 3.60   | 9.55± 3.97   |
| vae         | 2.01± 1.34   | 2.62± 1.25   | 3.12± 1.42   | 3.30± 1.41   | 4.19± 1.79   |

## Appendix G   Further results on Freeform Task

In Table 4, we report values on the freeform task using $L2$ distance inplace of Pearson correlation. The results are qualitatively similar, but there is overall less between-model variance in the values.

## Appendix H   Comparison with human data: Do models have compositional bias?

The multiple choice completion task from Section 5.2 in the main text was modeled after Experiment 1 in [31]. An intriguing result of that experiment was that peoples' responses are biased towards the CG completion. We asked whether any of the models shares this property. To do this, we measured the difference in accuracy when the prompt curve was sampled from the CG compared to when it was sampled from the SM. More precisely, let $\{y_0^i\}_i$ denote a collection of prompt curves, $\{y_1^i\}_i$ the corresponding completions generated by the CG, and $\{y_2^i\}_i$ the completions generated by the SM. Furthermore, define $z_i$ to be a binary variable that indicates whether $y_0^i$ was sampled from the CG or from the SM. In our setup, half of the prompt curves were sampled from the CG, meaning that $z_i$ assumes each of the two values with 50 percent probability. For a given model, the choice probabilites $\{(p_1^i, p_2^i)\}_i$ are given as in Section 5.2 in the main text. Then the accuracy difference is defined by

$$\Delta_{acc} := \mathbb{E}_i(p_1^i | z_i = CG) - \mathbb{E}_i(p_2^i | z_i = SM) \quad (4)$$

A short calculation shows that $\Delta_{acc}$ is directly related to the model's propensity to favor the CG completion over the SM completion:

$$\mathbb{E}_i(p_1^i) = \frac{1}{2}\mathbb{E}_i(p_1^i | z_i = CG) + \frac{1}{2}\mathbb{E}_i(p_1^i | z_i = SM) \quad (5)$$
$$= \frac{1}{2}\mathbb{E}(p_1^i | z_i = CG) + \frac{1}{2}(1 - \mathbb{E}_i(p_2^i | z_i = SM)) \quad (6)$$
$$= \frac{1}{2} + \frac{1}{2}\Delta_{acc} \quad (7)$$

In other words, $\Delta_{acc}$ is positive exactly when the CG completion is chosen more often than the SM completion. An unbiased model would have $\mathbb{E}_i(p_1^i) = \mathbb{P}_i(z_i = CG) = \frac{1}{2}$.

We see from Table 5 that all models had at least a weak form of the CG bias, in that they attained higher accuracy when the prompt was sampled from the CG. The contrastive model had the highest value of this bias, albeit the error bars overlap with the values for t-loss. However, all of these biases are smaller quantitatively than reported in Schulz, et. al. There it was found that people attain an accuracy of 32 percent when the prompt curve is from the SM (that is, they choose the CG completion 68 percent of the time), while they attain an accuracy of at least 71 percent when the prompt curve is from the CG. We say "at least", because in Schulz's experiment, there were actually three choices presented to the participants in the CG case: the CG completion, the SM completion, and an additional distractor completion. Thus we can estimate that, for people, the accuracy difference is given by

$$\Delta_{acc}^{people} \geq 39$$